\documentclass[letterpaper, 10 pt, conference]{ieeeconf}  

\IEEEoverridecommandlockouts                              

\usepackage{graphics} 
\usepackage[dvips]{graphicx}

\usepackage{color}

\usepackage{amsmath}
\usepackage{algorithm}
\usepackage{algpseudocode}

\usepackage{url}

\usepackage{tabularx, booktabs}
\newcolumntype{Y}{>{\centering\arraybackslash}X}

\usepackage[dvipsnames]{xcolor}

\newcommand{\argmin}[1]{\underset{#1}{\operatorname{arg}\,\operatorname{min}}\;}

\title{\LARGE \bf
Robot Imitation through Vision, Kinesthetic and Force Features\\with Online Adaptation to Changing Environments
}

\author{Raul Fernandez-Fernandez, Juan G. Victores, David Estevez and Carlos Balaguer
\thanks{All of the authors are members of the Robotics Lab
    research group within the Department of Systems Engineering and Automation,
    Universidad Carlos III de Madrid (UC3M). {\tt\small rauferna@ing.uc3m.es}}%
}

\begin{document}

\maketitle
\thispagestyle{empty}
\pagestyle{empty}

\begin{abstract}
Continuous Goal-Directed Actions (CGDA) is a robot imitation framework that encodes actions as the changes they produce on the environment. While it presents numerous advantages with respect to other robot imitation frameworks in terms of generalization and portability, final robot joint trajectories for the execution of actions are not necessarily encoded within the model. This is studied as an optimization problem, and the solution is computed through evolutionary algorithms in simulated environments. Evolutionary algorithms require a large number of evaluations, which had made the use of these algorithms in real world applications very challenging. This paper presents online evolutionary strategies, as a change of paradigm within CGDA execution. Online evolutionary strategies shift and merge motor execution into the planning loop. A concrete online evolutionary strategy, Online Evolved Trajectories (OET), is presented. OET drastically reduces computational times between motor executions, and enables working in real world dynamic environments and/or with human collaboration. Its performance has been measured against Full Trajectory Evolution (FTE) and Incrementally Evolved Trajectories (IET), obtaining the best overall results. Experimental evaluations are performed on the TEO full-sized humanoid robot with ``paint'' and ``iron'' actions that together involve vision, kinesthetic and force features.
\end{abstract}

\section{Introduction}

Robot imitation is a large area of study in robotics, which focuses on how a robot can learn an action based on user demonstrations.
Popular frameworks for robot imitation typically record the trajectories the robot or human performs during demonstrations,
successfully achieving reproduction of the average trajectory by the robot end-effector in Cartesian space.
The most prominent examples of these frameworks are Programming by Demonstration \cite{calinonlearning2007} and Dynamic Motion Primitives \cite{ijspeertdynamical2013}.
These algorithms shine for actions that are governed by geometry, such performing gestures in the air,
or performing simple manipulation tasks of moving an object from A to B.
Additional works have been performed to introduce active compliance \cite{Rozo2013} and obstacle avoidance \cite{Koert2016}.

However, a large body of actions that cannot be described solely in terms of human or robot geometric 
trajectories exists.
In addition to joint or Cartesian positions, visual and force features
provide relevant information 
when describing actions such as painting or ironing. 
Regarding visual features,
recent studies 
have  
focused on learning end-to-end mappings directly from raw images to the robot joint space \cite{Levine2016}.
These works can involve 
large sets of images for pre-training,
robot-environment physical interaction, and
additional hours 
for training. 

Continuous Goal-Directed Actions (CGDA) is a
feature-agnostic 
robot imitation framework \cite{moranteaction2014}.
Actions are encoded as time series of the variation of scalar features extracted from sensor data during user demonstrations.
While
this framework provides a rich infrastructure for generalizing actions,
this advantage comes at a cost.
Final robot joint or end-effector Cartesian trajectories are not necessarily encoded within the model.
Their components may have to be completely recomputed in order to comply with additional goals such as vision or force, or
may be discarded manually or through automatic feature selection algorithms \cite{moranteautomatic2015}.
This recomputation
is studied
as an optimization problem, and
has been 
solved through evolutionary algorithms in simulated environments.
CGDA requires no previous interaction with the environment, or additional training times. 

In this paper, the online evolutionary strategy paradigm for CGDA execution is presented.
The following contributions and consequences result from this
change of 
paradigm.

\begin{itemize}

\item Motor execution has been shifted and merged into the CGDA planning loop, enabling online adaptation for changing environments.
\item We demonstrate that the total time dedicated to mental simulation processes between motor executions is no longer dependent on the duration of the action.
\item The order of magnitude of results 
has been reduced from minutes with Incrementally
Evolved Trajectories (IET) \cite{morantehumanoid2015} to seconds with the presented 
online evolutionary strategy, Online Evolved Trajectories (OET).
\end{itemize}

\begin{figure}[h]
\centering 
\includegraphics[width=0.47 \textwidth]{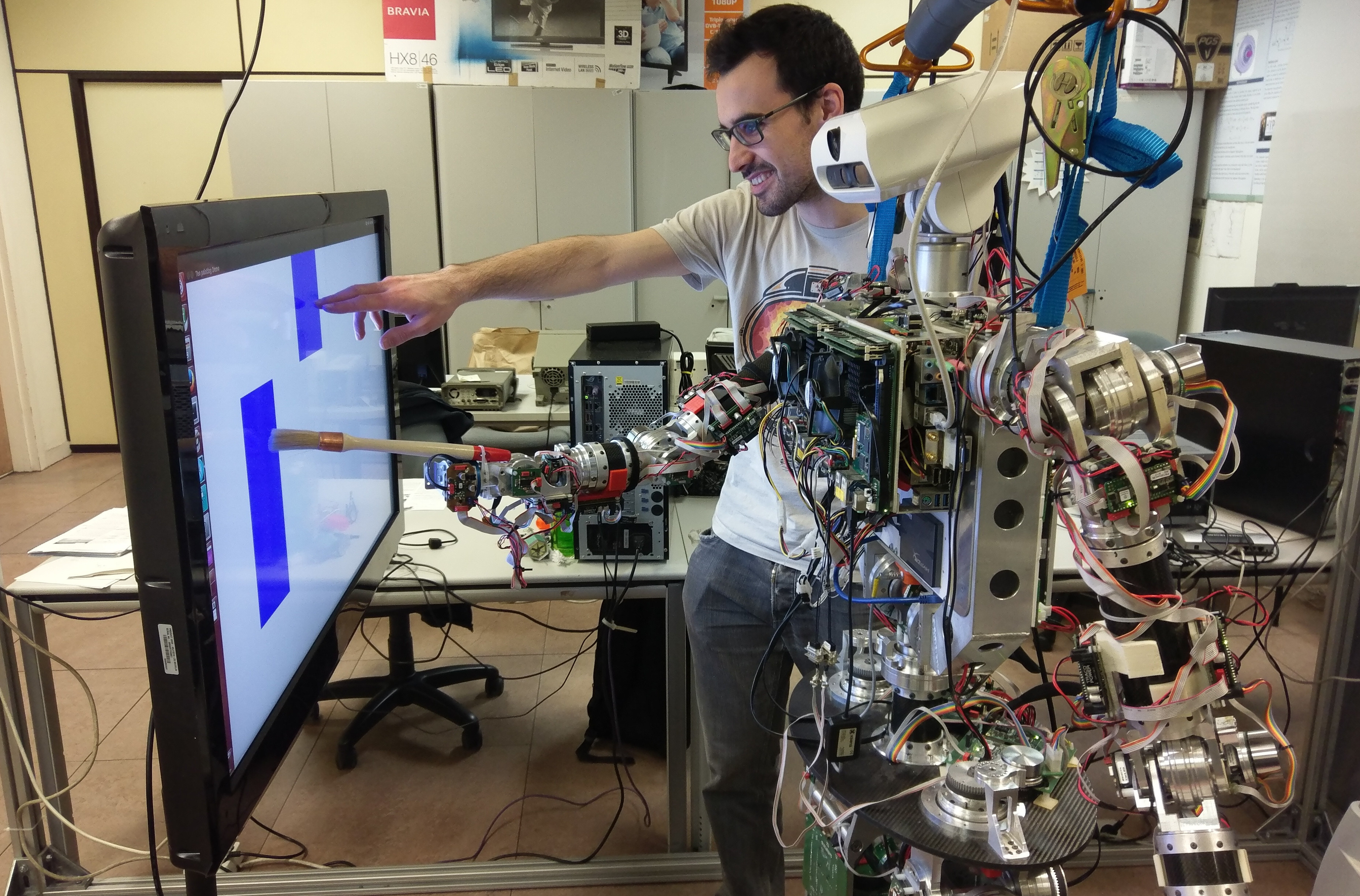}
\caption{Online Evolved Trajectories (OET) allows changes in the environment 
during execution,
as can be seen with the ``paint'' action.}
 \label{painting}
\end{figure}

The ``paint'' action, Figure \ref{painting}, has been used to 
evaluate OET for a pure visual feature action,
and an
``iron'' action was
used for kinesthetic and force features.

\section{CGDA Framework and Strategies}
\label{cgda}

CGDA is a framework for generalizing, recognizing and executing actions based on
scalar features extracted from sensor data. 
In CGDA, an action is modelled as a trajectory in a feature space of $m$ scalar features,
to represent the changes it produces on the environment.
Scalar features used in CGDA, in addition to the geometric trajectory of a specific robot or human configuration,
may include visual features of the environment, forces exerted by a given actuator, or even Cartesian positions
of moving objects in the environment.
Achieving
a state of the environment in which
the scalar features extracted from sensor perception match those of a given modelled action
is studied as an optimization problem in the execution stage.
The features are used as constraints to compute or recompute robot joint trajectories.

In CGDA not only the goal set of features, but also intermediate goals, must be achieved.
An action is sliced into $n$ intermediate goals,
computed as $n = \lfloor \frac{ D_{time} }{ T_{min} } \rfloor$,
where
$D_{time}$ is the average duration of user demonstrations, and
$T_{min}$ is the minimum 
time interval
between
intermediate goals.
The generalized representation of an action $X$ is
a trajectory in the $m$-dimensional
feature space with $n$ intermediate goals $X_j$ as defined in \eqref{feature_trajectory1}.

\begin{equation}\label{feature_trajectory1}
X=
\left(
\begin{array}{ccccc}
x_{00} & \cdots & x_{0j} & \cdots & x_{0n} \\
\vdots & \vdots & \vdots & \vdots & \vdots\\
x_{m0} & \cdots & x_{mj} & \cdots & x_{mn}
\end{array}
\right)
\end{equation}

Where $m$ may be set manually
or via 
feature selection algorithms 
\cite{moranteautomatic2015}.
Let $d_i$ be a discrete sample of any user demonstration for feature $i$,
then $x_{ij}$ is computed as in \eqref{feature_trajectory2}.

\begin{equation}\label{feature_trajectory2}
x_{ij} = \frac{1}{| d_i \in{[j,j+1]} |}\sum_{d_i \in{[j,j+1]}} d_i
\end{equation}

Recognition of an action is performed by comparing an observed action $O$ with a generalized action $X$. The discrepancy metric used is the sum
of costs of aligning each feature $i$ cost matrix $c_{p}(O_{i},X_{i})$.
Each cost matrix is computed as within Dynamic Time Warping \cite{Muller2007} 
as in \eqref{matrixcost}.

\begin{equation}\label{matrixcost}
c_{p}(O_{i},X_{i}) = \left( \begin{array}{ccc}
c(o_{i0},x_{i0}) & \cdots & c(o_{i0}, x_{in}) \\
\vdots & \vdots & \vdots \\
c(o_{in'},x_{i0}) & \cdots & c(o_{in'},x_{in}) \end{array} \right)
\end{equation}

Let $O_{time}$ be the duration of the observed action, $n'$ is computed as $n' = \lfloor \frac{ O_{time} }{ T } \rfloor$.

For execution, evolutionary algorithms are used to
compute robot joint trajectories.
Three different
strategies for CGDA execution have been
previously
proposed: 
Full Trajectory Evolution (FTE), Individual Evolution (IE), and Incrementally Evolved Trajectories (IET) \cite{morantehumanoid2015}.

In FTE, Algorithm \ref{Full},
each $individual$ of the population is composed by $DoF{\cdot}n$ parameters, where $DoF$ is the number of used degrees of freedom of the robot.
The full robot joint trajectory $U$ is generated 
attempting to reach all the intermediate goals simultaneously.
Execution and recognition are performed in an internal ``mental'' simulation,
where fitness $f$ is the
recognition discrepancy.
Termination conditions
are evaluated a maximum of $tc$ times,
while additionally monitoring
the evolution of $f$.

\begin{algorithm}
\caption{Full Trajectory Evolution (FTE)}\label{Full}
\begin{algorithmic}[1]
\Procedure{FTE}{$X$} 
\State $individuals\gets$initialize
\While{$\textbf{not}\ termination\_conditions$}
\For{\textbf{each} $individual$}
\State $U\gets$evolve($DoF{\cdot}n$)
\State $O\gets$mental\_execution($U$)
\State $f\gets$mental\_recognition($O,X$)
\EndFor
\EndWhile
\State motor\_execution($U$)
\EndProcedure
\end{algorithmic}
\end{algorithm}

In IE, Algorithm \ref{Individual},
each $individual$
is composed by $DoF$ parameters.
Joint positions $U_{j}$ are generated independently for each
intermediate goal.

\begin{algorithm}
\caption{Individual Evolution (IE)}\label{Individual}
\begin{algorithmic}[1]
\Procedure{IE}{$X$}
\State $individuals\gets$initialize
\For{$j < n$}
\While{$\textbf{not}\ termination\_conditions$}
\For{\textbf{each} $individual$}
\State $U_{j}\gets$evolve($DoF$)
\State $O_{j}\gets$mental\_execution($U_{j}$)
\State $f\gets$mental\_recognition($O_{j},X_{j}$)
\EndFor
\EndWhile
\EndFor
\State motor\_execution($U$)
\EndProcedure
\end{algorithmic}
\end{algorithm}

In IET, Algorithm \ref{IET},
joint positions 
$U_{j}$ are 
generated 
for each
intermediate goal
after the mental execution of $U_{[0,j-1]}$.

\begin{algorithm}[h!]
\caption{Incrementally Evolved Trajectories (IET)}\label{IET}
\begin{algorithmic}[1]
\Procedure{IET}{$X$}
\State $individuals\gets$initialize
\For{$j < n$}
\While{$\textbf{not}\ termination\_conditions$}
\For{\textbf{each} $individual$}
\State mental\_execution($U_{[0,j-1]}$)
\State $U_{j}\gets$evolve($DoF$)
\State $O_{j}\gets$mental\_execution($U_{j}$)
\State $f\gets$mental\_recognition($O_{j},X_{j}$)
\EndFor
\EndWhile
\EndFor
\State motor\_execution($U$)
\EndProcedure
\end{algorithmic}
\end{algorithm}

Experimental evidence from previous publications has determined FTE to be the
strategy that requires most evaluations for fitness convergence \cite{morantehumanoid2015}.
The main intuition behind this
large amount of required evaluations
is that evolutionary algorithms are greatly affected by the
size of the
search space.
In FTE,
the search space is $(DoF{\cdot}n)$-dimensional, which is proportional to the number of intermediate goals.

IE is the strategy that requires least evaluations for fitness convergence
of the three presented strategies, with a $DoF$-dimensional search space.
However, joint positions are generated independently for each intermediate goal,
which  
leads to 
an inherent issue.
In the case of final intermediate goals,
this means accomplishing the majority of a final goal 
with a single robot joint position. 
Let a ``paint'' action be the use case,
accomplishing this
is not realistic.
Fitness convergence
may result 
in the same robot joint position
for two or more different intermediate goals.
This is not only a \textit{duplicate} effect, but also represents a
step loss
or loss of time
to achieve a different goal
contributing to the general solution.

In IET, the robot joint trajectory that has been computed to
achieve the previous intermediate goals is 
executed in the simulation
before generating each new robot joint position.
This provides
awareness 
of 
the
previously achieved intermediate goals,
avoiding
the inherent issue
described for IE.
The search space is $DoF$-dimensional as in IE.

\section{Online Evolutionary Strategies}
\label{oes}

In
the previously presented CGDA execution strategies,
there was 
a mental 
process
of
execution and recognition 
in a simulated environment
while monitoring fitness evolution,
and finally motor execution was performed.
In this sense, they can be considered offline planning algorithms.
The general layout of an offline CGDA execution evolutionary strategy
is summarized 
in Algorithm \ref{offline}, where
planning termination conditions encompass all the loop conditions.

\begin{algorithm}
\caption{Offline Evolutionary Strategy}\label{offline}
\begin{algorithmic}[1]
\Procedure{Offline}{}
\While{$\textbf{not}\ planning\_termination\_conditions$}
\State mental\_process\_loop
\EndWhile
\State motor\_execution($U$)
\EndProcedure
\end{algorithmic}
\end{algorithm}

This paper presents a new  
layout for CGDA execution evolutionary strategies,
namely online evolutionary strategies.
The general layout of an online CGDA execution evolutionary strategy
is summarized 
in Algorithm \ref{online}.

\begin{algorithm}
\caption{Online Evolutionary Strategy} \label{online}
\begin{algorithmic}[1]
\Procedure{Online}{}
\While{$\textbf{not}\ planning\_termination\_conditions$}
\State mental\_process\_loop
\State motor\_execution($U_{j}$)
\EndWhile
\EndProcedure
\end{algorithmic}
\end{algorithm}

Motor execution in
online Algorithm \ref{online} is
of individual motor movements $U_j$, rather than
the full robot joint trajectory $U$ of offline Algorithm \ref{offline}.
This motor execution of $U_j$ is performed:

\begin{enumerate}
\item Once per intermediate goal.
\item After a single mental process loop.
\end{enumerate}

The consequences are, respectively:

\begin{enumerate}
\item Movements should occur $n$ times.
\item The repetitions of mental process loops between motor executions is reduced by a factor of $n$.
\end{enumerate}

A further consequence of 
(2) 
is that the total time dedicated to mental processes between motor executions
is no longer dependent of $n$, and is therefore independent of the duration of the action.

\section{The OET Algorithm}
\label{oet}

Online Evolved Trajectories (OET) is presented in this paper as an evolutionary strategy to effectively 
reduce computation times for execution
inside the CGDA framework for real world applications. 
OET 
is a concrete implementation of
an online 
evolutionary 
strategy for real world applications within the CGDA framework. 
The pseudcode of this strategy is presented in Algorithm \ref{OET}. 

\begin{algorithm}
\caption{Online Evolved Trajectories (OET)} \label{OET}
\begin{algorithmic}[1]
\Procedure{OET}{$X$}
\State $individuals\gets$initialize
\While{$\textbf{not}\ oet\_termination\_conditions$}
   \State $P_{t}\gets$sensor\_perception
   \State $j\gets$localization($P_{t}$)
   \While{$\textbf{not}\ termination\_conditions$}
      \For{\textbf{each} $individual$}
      \State $U_{j+1}\gets$evolve($DoF$)
      \State $O_{j+1}\gets$mental\_execution($U_{j+1}$)
      \State $f\gets$mental\_recognition($O_{j+1},X_{j+1}$)
      \EndFor
   \EndWhile
   \State motor\_execution($U_{j+1}$)
\EndWhile
\EndProcedure
\end{algorithmic}
\end{algorithm}

OET termination conditions are evaluated a maximum of $otc$ times,
while additionally monitoring that the final goal has not been achieved.
To achieve 
introducing motor execution within the planning algorithm loop,
real world sensor perception and localization steps
are 
additionally performed.

\subsection{Sensor Perception}
In the Sensor Perception step, 
the system extracts the 
scalar features from the 
real world environment sensor data. 
An updated vector $P_t$ in the $m$-dimensional feature space of $X$ is obtained from the current state of the world, at time $t$, as in \eqref{world_feature}. 

\begin{equation}\label{world_feature}
P_{t}=[p_{0t}, p_{1t}, p_{2t}, p_{3t}, p_{4t}..., p_{mt}]^T
\end{equation}

\subsection{Localization}

In the Localization step, the features extracted from the Sensor Perception step are used to locate the intermediate goal that corresponds with the current environment state. The objective of this step is to find the intermediate goal $j$ of the feature trajectory $X$ that reduces the discrepancy between $P_{t}$ and $X_{j}$ as in equation \eqref{time_localization}.

\begin{equation}\label{time_localization}
j=\argmin{j\in[j_{prev},n]}( \|P_{t}-X_j\|_p) 
\end{equation}

Where $j_{prev}$ is the index of the previously accomplished intermediate goal, and $p$ is the order of the norm used for Localization, preferably the Euclidean L2 norm.

\section{Experiments}
\label{experiments}

Three different evolutionary strategies for CGDA were tested in the experiments of this paper: Full Trajectory Evolution (FTE), Incrementally Evolved Trajectories (IET), and the Online Evolved Trajectories (OET).
Individual Evolution (IE) was not used due to the 
inherent issue explained at the end of Section \ref{cgda}.
The actions chosen for the experiments were the ``paint'' and ``iron'' actions,
as use cases that together include relevant
visual, kinesthetic and force features.

The robotic platform used 
was TEO, a 
full-sized humanoid robot
\cite{TEO}. 
For demonstrations of the ``paint'' action, a paintbrush was attached to the left end-effector of the robot, and the 6 degrees of freedom of the left arm in gravity compensation mode were used. 
An ASUS Xtion PRO LIVE RGB-D was used to extract the percentage of painted wall. 
For the ``iron'' action demonstrations, an iron was installed as the right end-effector using custom 3D printed parts, and the 6 degrees of freedom of the right arm in gravity compensation mode were used.
The CUI absolute encoders present in each of the joints of the robot were used to obtain the Cartesian position of the end-effector via forward kinematics.  ``iron'' action.
Finally, a 
JR3 force/torque sensor equipped in the right wrist of the robot was used to measure force features in the ``iron'' demonstrations.
For all of the execution strategies, 3 of the 6 degrees of freedom of the right arm of the robot were used for the evolution, keeping all the other joints (including torso, legs and head) static.

ECF \cite{picek2011evaluation}
was used 
as the C++ framework for
evolutionary computation.
YARP \cite{fitzpatrick2008towards} was used for internal and robot component communications.
OpenRAVE \cite{diankovthesis} was used for the simulation environment.
The
experimental datasets 
and presented CGDA
strategies
have been
open-sourced\footnote{\url{https://github.com/roboticslab-uc3m/xgnitive}}.

Steady State Tournament (SST)
has been the standard
evolutionary algorithm
used in CGDA implementations,
and has
also been used in the experiments in this paper.
The presented strategies are situated 
a layer above evolutionary algorithms such as SST, which can be considered a
back-end.
Their comparison
should not be 
affected by the selection of a specific set of back-end shared parameters.
Parameters have been set to achieve reasonable 
execution 
times on a single core of a single machine.

Following this assumption,
the SST parameters for all the strategies were set to
a population of 10 individuals,  
a tournament size of 3 individuals,
and an individual mutation probability of 60\%.
The search space of each individual was 
bounded 
between 
-15 and 100, which corresponds to the individual robot arm joint limits expressed in degrees.
 
FTE termination conditions were to reach a zero fitness $f$ value,
maximum 
$tc=300$,
or maximum $tc$ without improvement in fitness $tcf=75$.
For IET, 
$tc$ and $tcf$ are scaled by $n$ due to the outer $n$ loop, resulting in $tc=300/n$ and $tcf=75/n$.
Finally, for OET, $tc$ and $tcf$ are scaled by $otc$ due to the outer $otc$ loop, resulting in $tc=300/otc$ and $tcf=75/otc$.

The following metrics were used within the development of the experiments:
\begin{itemize}
\item Evaluations: The total number of passes through mental recognition.
\item Discrepancy: The final achieved fitness $f$.
\item Real Iteration Time ($RIT$): Time
between two contiguous motor executions, as defined in \eqref{world_frequency}.

\begin{equation}\label{world_frequency}
RIT=t_j-t_{j_{prev}}
\end{equation}
\end{itemize}

\subsection{Paint}

The ``paint'' action is a representative use case presented in previous work of the authors \cite{morantehumanoid2015}.
While in previous work the generalized ``paint'' action was generated synthetically as a linear growth from 0\% to 100\%
of the painted portion of a tracked object (a wall), this feature trajectory was now generated from 4 user demonstrations. 

\begin{figure} [htpb]
 \begin{center}
    \includegraphics[width=0.46\textwidth]{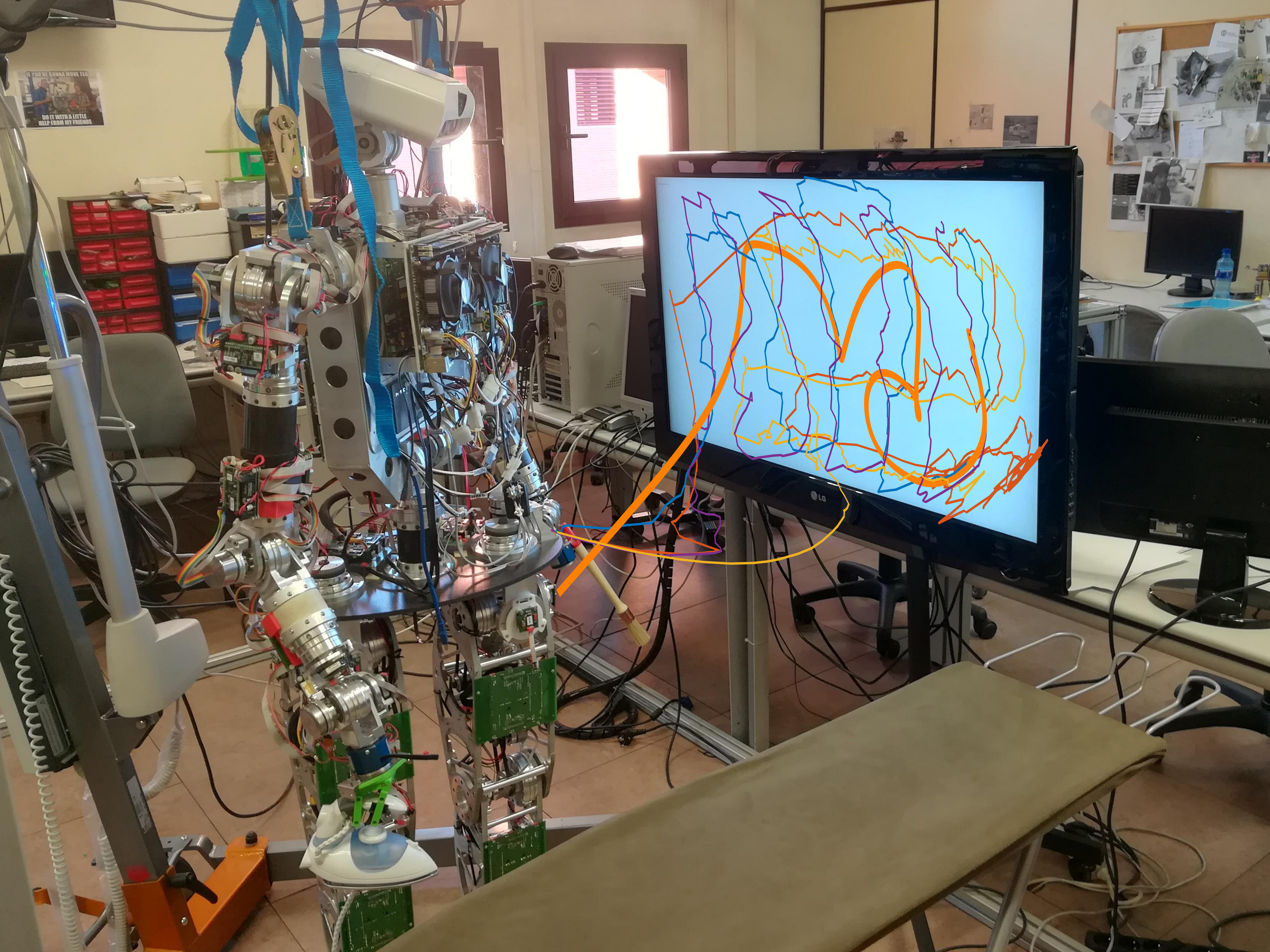}
\caption{ 
The ``paint'' action demonstrations.  
The additional 
thick line depicts
a failed
pure
geometrical approach, with $K=7$ and $T=600$ as in \cite{calinonlearning2007}.
}
 \label{paintGMM}
 \end{center}
\end{figure}

Each of the demonstrations was deliberatively performed following a different geometrical trajectory, as depicted in Figure \ref{paintGMM}.
The figure 
also depicts the geometrical model generated using Gaussian Mixture Models and Gaussian Mixture Regression as in \cite{calinonlearning2007}. The method
achieves painting 43.75\% of the surface, as
the mixture of different geometrical trajectories
results in a trajectory similar to their average,
which may be or not relevant for performing the action.

\begin{table*}[thpb]
\vspace{1em}
\caption{Experimental results for the ``paint'' action using the three strategies presented in this paper}
\label{exp_res}
\centering
\begin{tabularx}{\textwidth}{|c *{9}{|Y}|}
	\cline{2-9}
	 \multicolumn{1}{c|}{} & \multicolumn{2}{c|}{Evaluations} & \multicolumn{2}{c|}{Discrepancy ($f$)} & \multicolumn{2}{c|}{ Real Iteration Time ($RIT$) [s]} & \multicolumn{2}{c|}{ Painted Wall [\%]} \\ \hline
	 Strategy & $\mu$ & $\sigma$ & $\mu$ & $\sigma$ & $\mu$ & $\sigma$ & $\mu$ & $\sigma$  \\ \hline
	 FTE & 1716 &    231.80 & 49.48 & 7.40  & 272.3 & 68.48 & 85.4  & 3.6  \\ \hline
	 IET & 1153 &    161.65 & 54    & 25.36 & 143   & 25.87   & 72.9  & 15.72 \\ \hline
	 OET & 1603.33 & 20.82  & 40.19 & 3     & 4     & 0.6    & 89.58 & 3.6 \\ \hline
	 
	\end{tabularx}
\end{table*}

The average demonstration time of the ``paint'' action was $D_{time}=130.2\ s$.
Selecting a low $T_{min}$ would result in an intractable value of $n$ for FTE, due to the $DoF\cdot n$ size of its search space.
$T_{min}=10\ s$ was set, resulting in $n=13$ intermediate goals for comparison of the strategies.

\begin{figure} [thpb]
 \begin{center}
    \includegraphics[width=0.5 \textwidth]{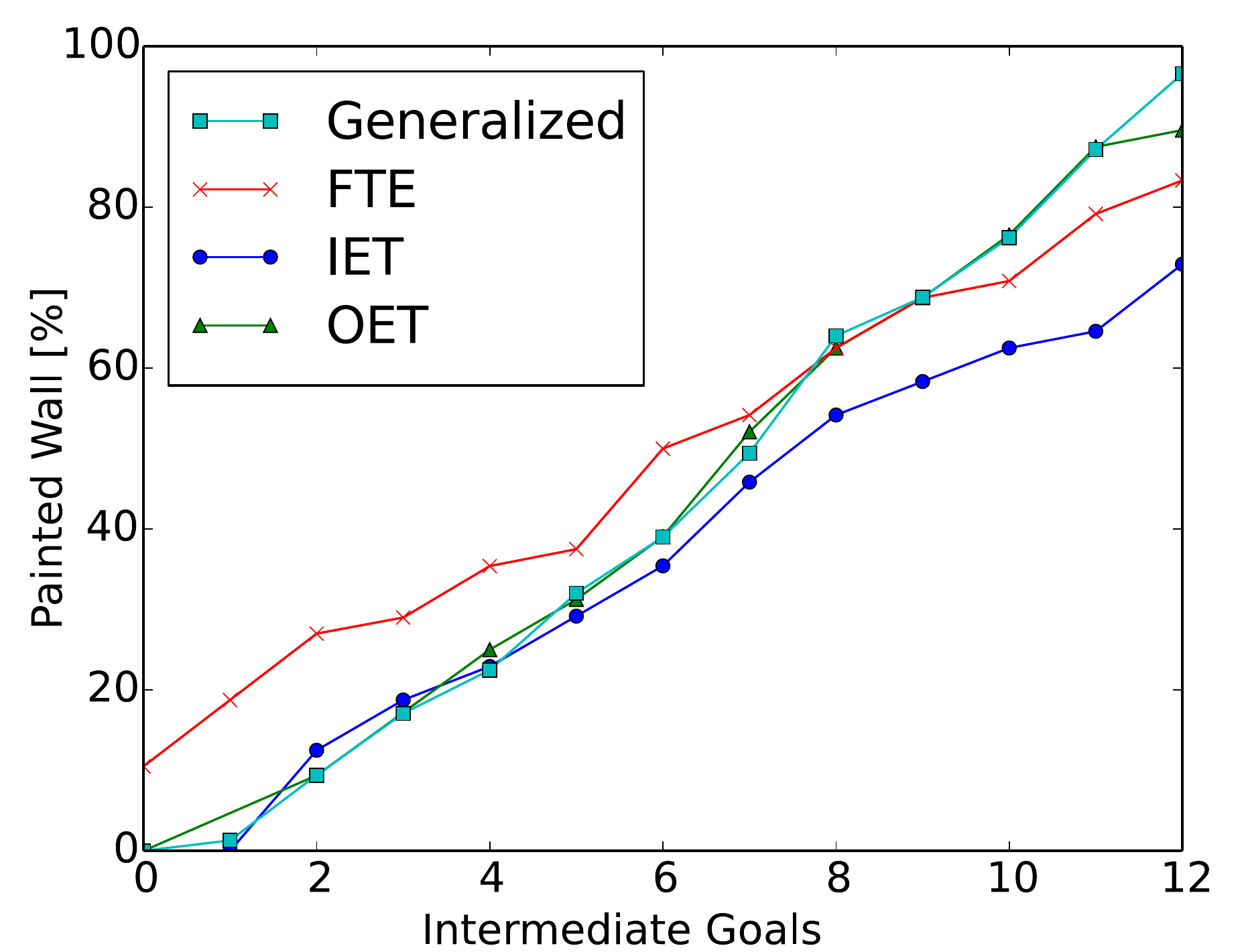}
    \caption{ Generalized action obtained from user demonstrations 
compared to
the intermediate goals achieved
by each of the strategies.} 
 \label{Trajectories}
 \end{center}
\end{figure}

The results obtained from the CGDA execution strategy experiments for ``paint'' are shown in Table \ref{exp_res}, where averages and standard deviations were extracted from 3 repetitions of each experiment.
Figure \ref{Trajectories} shows a comparison
of the achievement of intermediate goals
with each of the strategies,
compared to the reference generalized action obtained from the user demonstrations.

Similar to previous experimental evidence \cite{morantehumanoid2015}, FTE was the strategy that took most evaluations to converge, as a result of the size of the search space. 
Discrepancy was not the highest, 
despite 
the apparent lack of correlation 
with respect to
the generalized action in Figure \ref{Trajectories}. 
This is 
due to the Dynamic Time Warping metric used in mental recognition.
FTE is also the slowest strategy in terms of $RIT$,
accounting for all the evaluations before motor execution.

IET requires 
less evaluations
and $RIT$ than 
FTE, as a result of the reduced search space.
However, 
IET has a larger discrepancy and achieves a lower percentage of painted wall than FTE.
This is because IET may suffer the effects of non-optimal decisions 
for initial intermediate goals.

OET 
results in more evaluations than IET in Table \ref{exp_res}, as 
OET may
perform diffferent motor executions until it achieves an
intermediate goal.
Figure \ref{Trajectories}
is a compact representation that 
depicts the percentage of painted wall after achieving each intermediate goal.
OET obtained the best result in terms of $RIT$, with an average of 4 seconds between real motor executions.
Its final achieved percentage of painted wall is also the highest, and it additionally minimizes discrepancy.

\begin{table*}[thpb]
\vspace{1em}
\caption{Experimental results for the ``iron'' action using the three strategies presented in this paper}
\label{exp_res_iron}
\centering
\begin{tabularx}{\textwidth}{|c *{7}{|Y}|}
	\cline{2-7}
	 \multicolumn{1}{c|}{} & \multicolumn{2}{c|}{Evaluations} & \multicolumn{2}{c|}{Discrepancy ($f$)} & \multicolumn{2}{c|}{ Real Iteration Time ($RIT$) [s]} \\ \hline
	 Strategy & $\mu$ & $\sigma$ & $\mu$ & $\sigma$ & $\mu$ & $\sigma$ \\ \hline
FTE & 3010 & 0      & 0.7  & 0.09 & 2481 & 1.73 \\ \hline
IET & 1588 & 113.74 & 0.59 & 0.05 & 30.30 & 2.69 \\ \hline
OET & 1010 & 400.37 & 0.30 & 0.07 & 1.44 & 0.16 \\ \hline
	\end{tabularx}
\end{table*}

\subsection{Iron}

\begin{figure} [h]
 \begin{center}
     \includegraphics[width=0.47\textwidth]{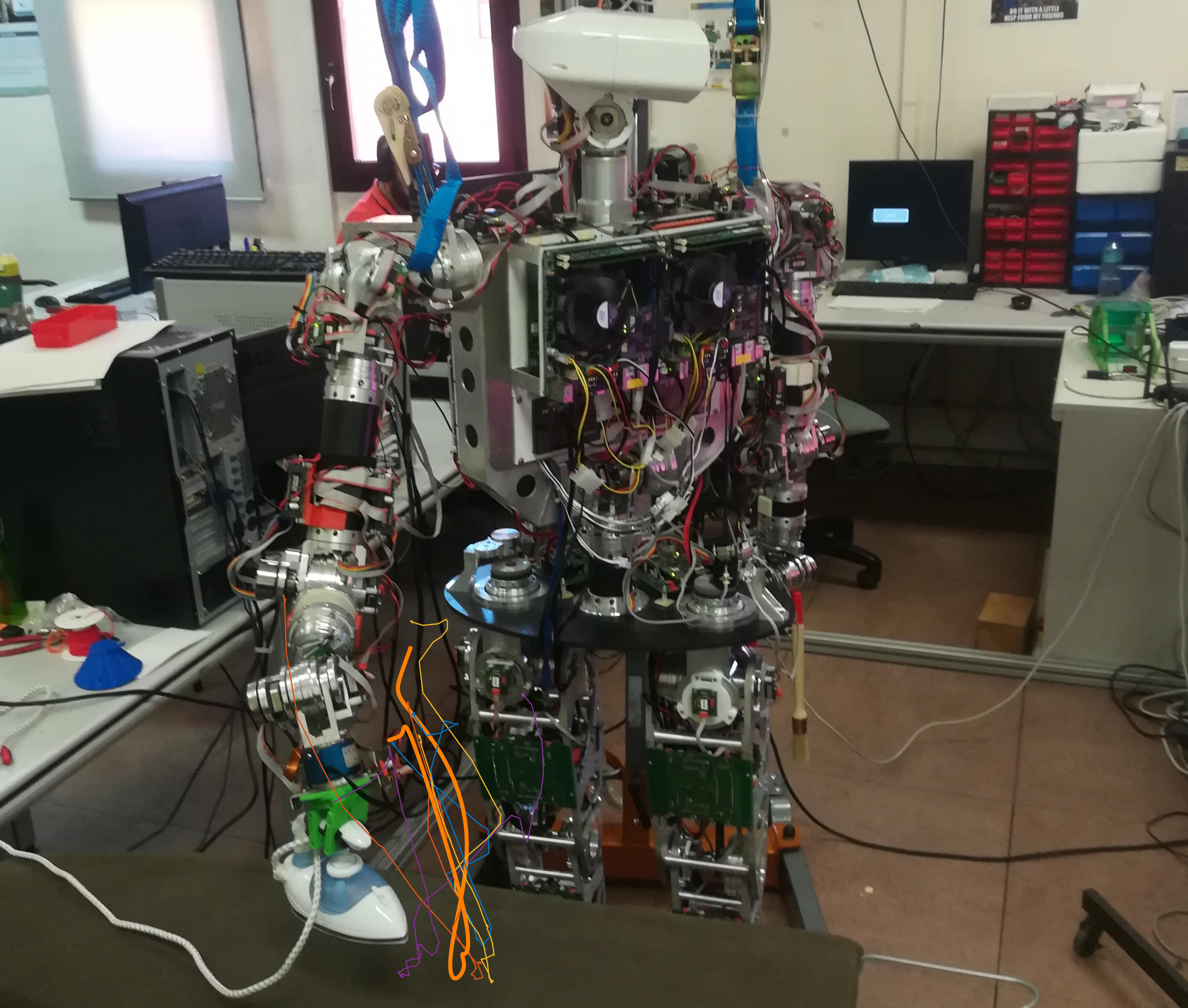}
\caption{
The ``iron'' action demonstrations.
The additional thick line depicts
a failed
pure
geometrical approach, with $K=5$ and $T=150$ as in \cite{calinonlearning2007}.
}
 \label{iron3d}
 \end{center}
\end{figure}

The generalized action for the `iron' action was generated from 4 demonstrations, depicted in Figure \ref{iron3d}.
The relevant features in this action were the end-effector Cartesian positions and the force exerted by the iron
measured on its vertical axis.
The objective was to descend on the ironing board, apply 30 N force, and then ascend again.

The figure also depicts the pure geometrical model generated using Gaussian Mixture Models and Gaussian Mixture Regression as in \cite{calinonlearning2007}.
In this case, while geometrically accurate, the measured force was close to zero.

The average demonstration time of the ``iron'' action was $D_{time}=28.1\ s$.
$T_{min}=3\ s$ was set, resulting in $n = 9$ intermediate goals for comparison of the strategies.
The results obtained from the experiments are shown in Table \ref{exp_res_iron},
extracted from 3 repetitions of each experiment.

For the ``iron'' action, FTE took the maximum amount of evaluations possible, composed by the initialization of the 10 individuals, and reaching $tc=300$ 
with this population.
FTE discrepancy and $RIT$ were also the highest for this action,
while 
intermediate results were obtained with IET.

OET obtained the best overall results for the ``iron'' action.
The 
$RIT$ average 1.4 second mark is similar to the times of human mental simulations as measured in \cite{Hamrick2015}. 

\section{Conclusions}
\label{conclusions}

A change of paradigm in evolutionary strategies for CGDA execution,
from offline to online evolutionary strategies, is presented in this paper.
Previously developed algorithms for CGDA execution subscribed to a model where
planning was performed in mental simulations, and the final computed trajectory was sent
to the robot for motor execution.
Online evolutionary strategies
reduce the time dedicated to mental processes between motor executions by
shifting motor execution into the planning loop.

A concrete implementation of
an online evolutionary strategy, Online Evolved Trajectories (OET),
has additionally been introduced.
OET is 
an 
online evolutionary strategy
for CGDA in real world applications, including dynamic environments or human intervention/collaboration. 
It 
enables human interventions
similar to the pure geometric approach of \cite{Vogt2016}, 
enhanced by complementary features such as vision and force.
These features are recorded simultaneously and agnostically.
This is an improvement to previous literature on robot imitation of an ``iron'' action \cite{kormushev2011imitation},
where geometrical trajectories are learned first, and then forces are demonstrated using a separate haptic device during the execution of the previously learned geometrical trajectory.
The results obtained show a notable improvement over the previous offline strategies used by the authors, experiencing large improvements 
not only in terms of
elapsed time
between motor executions, but also in terms of
overall fitness of the Continuous Goal-Directed Action. 

OET has opened a new range of possible real world applications to the CGDA framework. The implementation of real world actions, where the environment experiences external changes, or collaborative tasks where the user helps the robot to perform the action, is now feasible within the CGDA framework.

Future lines of research include reducing $RIT$, for instance through the use of parallelism.
As $RIT$ is minimized, adaptive rates of $T_{min}$ can additionally be incorporated.

\section{ACKNOWLEDGMENT}

The research leading to these results has received funding from the RoboCity2030-III-CM project (Rob\'otica aplicada a la mejora de la calidad de vida de los ciudadanos. Fase III; S2013/MIT-2748), funded by Programas de Actividades I+D en la Comunidad de Madrid and cofunded by Structural Funds of the EU.




\bibliographystyle{IEEEtran}
\bibliography{iros2018.bib}

\begin{thebibliography}{10}
\providecommand{\url}[1]{#1}
\csname url@samestyle\endcsname
\providecommand{\newblock}{\relax}
\providecommand{\bibinfo}[2]{#2}
\providecommand{\BIBentrySTDinterwordspacing}{\spaceskip=0pt\relax}
\providecommand{\BIBentryALTinterwordstretchfactor}{4}
\providecommand{\BIBentryALTinterwordspacing}{\spaceskip=\fontdimen2\font plus
\BIBentryALTinterwordstretchfactor\fontdimen3\font minus
  \fontdimen4\font\relax}
\providecommand{\BIBforeignlanguage}[2]{{%
\expandafter\ifx\csname l@#1\endcsname\relax
\typeout{** WARNING: IEEEtran.bst: No hyphenation pattern has been}%
\typeout{** loaded for the language `#1'. Using the pattern for}%
\typeout{** the default language instead.}%
\else
\language=\csname l@#1\endcsname
\fi
#2}}
\providecommand{\BIBdecl}{\relax}
\BIBdecl

\bibitem{calinonlearning2007}
S.~Calinon, F.~Guenter, and A.~Billard, ``On {Learning}, {Representing}, and
  {Generalizing} a {Task} in a {Humanoid} {Robot},'' \emph{IEEE Transactions on
  Systems, Man, and Cybernetics, Part B (Cybernetics)}, vol.~37, no.~2, pp.
  286--298, Apr. 2007.

\bibitem{ijspeertdynamical2013}
A.~J. Ijspeert, J.~Nakanishi, H.~Hoffmann, P.~Pastor, and S.~Schaal,
  ``\BIBforeignlanguage{eng}{Dynamical movement primitives: learning attractor
  models for motor behaviors},'' \emph{\BIBforeignlanguage{eng}{Neural
  Computation}}, vol.~25, no.~2, pp. 328--373, Feb. 2013.

\bibitem{Rozo2013}
L.~Rozo, S.~Calinon, D.~G. Caldwell, P.~Jimenez, and C.~Torras, ``Learning
  collaborative impedance-based robot behaviors,'' in \emph{{AAAI} Conference
  on Artificial Intelligence}, Bellevue, WA, USA, 2013, pp. 1422--1428.

\bibitem{Koert2016}
D.~Koert, G.~Maeda, R.~Lioutikov, G.~Neumann, and J.~Peters, ``Demonstration
  based trajectory optimization for generalizable robot motions,'' in
  \emph{Proceedings of the International Conference on Humanoid Robots
  (HUMANOIDS)}, 2016.

\bibitem{Levine2016}
S.~Levine, C.~Finn, T.~Darrell, and P.~Abbeel, ``End-to-end training of deep
  visuomotor policies,'' \emph{Journal of Machine Learning Research}, vol.~17,
  no.~39, pp. 1--40, 2016.

\bibitem{moranteaction2014}
S.~Morante, J.~G. Victores, A.~Jard\'on, and C.~Balaguer, ``Action effect
  generalization, recognition and execution through continuous goal-directed
  actions,'' in \emph{2014 IEEE International Conference on Robotics and
  Automation (ICRA)}.\hskip 1em plus 0.5em minus 0.4em\relax {IEEE}, 2014, pp.
  1822--1827.

\bibitem{moranteautomatic2015}
S.~Morante, J.~G. Victores, and C.~Balaguer, ``Automatic demonstration and
  feature selection for robot learning,'' in \emph{2015 IEEE-RAS 15th
  International Conference on Humanoid Robots (Humanoids)}.\hskip 1em plus
  0.5em minus 0.4em\relax IEEE, Nov. 2015, pp. 428--433.

\bibitem{morantehumanoid2015}
S.~Morante, J.~G. Victores, A.~Jard\'on, and C.~Balaguer, ``Humanoid robot
  imitation through continuous goal-directed actions: an evolutionary
  approach,'' \emph{Advanced Robotics}, vol.~29, no.~5, pp. 303--314, 2015.

\bibitem{Muller2007}
M.~M\"uller, \emph{Dynamic Time Warping}.\hskip 1em plus 0.5em minus
  0.4em\relax Berlin, Heidelberg: Springer Berlin Heidelberg, 2007, pp. 69--84.

\bibitem{TEO}
S.~Mart\'inez, C.~A. Monje, A.~Jard\'on, P.~Pierro, C.~Balaguer, and D.~Munoz,
  ``Teo: Full-size humanoid robot design powered by a fuel cell system,''
  \emph{Cybernetics and Systems}, vol.~43, no.~3, pp. 163--180, 2012.

\bibitem{picek2011evaluation}
S.~Picek, M.~Golub, and D.~Jakobovic, ``Evaluation of crossover operator
  performance in genetic algorithms with binary representation,'' in
  \emph{International Conference on Intelligent Computing}.\hskip 1em plus
  0.5em minus 0.4em\relax Springer, 2011, pp. 223--230.

\bibitem{fitzpatrick2008towards}
P.~Fitzpatrick, G.~Metta, and L.~Natale, ``Towards long-lived robot genes,''
  \emph{Robotics and Autonomous systems}, vol.~56, no.~1, pp. 29--45, 2008.

\bibitem{diankovthesis}
R.~Diankov, ``Automated construction of robotic manipulation programs,'' Ph.D.
  dissertation, Carnegie Mellon University, Robotics Institute, August 2010.

\bibitem{Hamrick2015}
J.~B. Hamrick, K.~A. Smith, T.~L. Griffiths, and E.~Vul, ``{Think again? The
  amount of mental simulation tracks uncertainty in the outcome},'' \emph{37th
  Annual Conference of the Cognitive Science Society}, vol.~1, 2015.

\bibitem{Vogt2016}
D.~Vogt, S.~Stepputtis, R.~Weinhold, B.~Jung, and H.~B. Amor, ``{Learning
  Human-Robot Interactions from Human-Human Demonstrations ( with Applications
  in Lego Rocket Assembly )},'' \emph{2016 IEEE-RAS International Conference on
  Humanoid Robots (Humanoids 2016)}, pp. 142--143, 2016.

\bibitem{kormushev2011imitation}
P.~Kormushev, S.~Calinon, and D.~G. Caldwell, ``Imitation learning of
  positional and force skills demonstrated via kinesthetic teaching and haptic
  input,'' \emph{Advanced Robotics}, vol.~25, no.~5, pp. 581--603, 2011.

\end{thebibliography}

\end{document}